\pdfoutput=1
\documentclass[11pt]{article}

\usepackage{acl}
\usepackage{booktabs}
\usepackage{array}
\usepackage{times}
\usepackage{latexsym}
\usepackage{subfigure}
\usepackage{float}
\usepackage{amsmath}
\usepackage{algorithm,algpseudocode}
\usepackage[T1]{fontenc}
\usepackage{footnote}

\usepackage[utf8]{inputenc}
\usepackage{multirow}
\usepackage{microtype}
\usepackage{graphicx}
\newcommand{\modelname}{\texttt{InstructionSpeak}}
\newcommand{\taskname}{\texttt{ConTinTin}}

\newcommand{\kashabidata}{\textsc{Natural-Instructions}}
\newcommand{\historytraining}{\textsc{History Training}}
\newcommand{\negativetraining}{\textsc{Negative Training}}

\title{\taskname: Continual Learning from Task Instructions}

\author{Wenpeng Yin\thanks{Work was done at Salesforce Research.} \\ Temple University \\ wenpeng.yin@temple.edu
        \And  
        Jia Li \\ Salesforce Research \\jia.li@salesforce.com
        \And
        Caiming Xiong \\ Salesforce Research \\ cxiong@salesforce.com}
\begin{document}
\maketitle
\begin{abstract}
The mainstream machine learning paradigms for NLP often work with two underlying presumptions. First, the target task is predefined and static; a system merely needs to learn to solve it exclusively. Second, the supervision of a task mainly comes from a set of labeled examples. A question arises: how to build a system that can keep learning new tasks from their instructions?

This work defines a new learning paradigm \taskname\enspace (\textbf{Contin}ual Learning from \textbf{T}ask \textbf{In}structions), in which a system should learn a sequence of new tasks one by one, each task is explained by a piece of textual instruction. The system is required to (i) generate the expected outputs of a new task by learning from its instruction, (ii) transfer the  knowledge acquired from upstream tasks to help solve downstream tasks (i.e., forward-transfer), and (iii)  retain or even improve the performance on  earlier tasks after learning  new tasks (i.e., backward-transfer). This new problem is studied on a stream of more than 60 tasks, each equipped with an instruction. Technically, our method \modelname~contains two strategies that make full use of task instructions to improve forward-transfer and backward-transfer: one is to learn from negative outputs, the other is to re-visit instructions of previous tasks. To our knowledge, this is the first time to study \taskname~in NLP. In addition to the problem formulation and our promising approach, this work also contributes to providing rich analyses  for the community to  better understand this novel learning problem.

\end{abstract}

\section{Introduction}

The main goal of machine learning algorithms lies in seeking   supervision for solving a target task. Traditionally, the supervision is extracted from a set of labeled examples. The learner constructs a decision function that generalizes beyond the  seen examples. While this paradigm has been tremendously successful for many NLP problems, an inherent drawback exists in it: the learner can only be as good as the provided data \cite{DBLPldwasserR14}. Learning, therefore, relies on annotating a large volume of training data, an expensive and time-consuming process. To alleviate the costly demand for task-specific annotation (referred as $\mathcal{S}_0$ hereafter), the human learning process suggests at least two sources of alternative supervision: one is to accumulate knowledge from tasks learned in the past ($\mathcal{S}_1$) \cite{montague1970universal,DBLPThrunM95,chomsky2009syntactic};  the other is to learn from natural instructions ($\mathcal{S}_2$)   describing a high-level story about  target tasks  \cite{DBLPldwasserR14}. Unfortunately, we rarely see the joint power of $\mathcal{S}_1$ and $\mathcal{S}_2$.

In this work, we present a new learning paradigm \taskname~ -- \textbf{contin}ual Learning from \textbf{t}ask   \textbf{in}structions. In \taskname, each task is given an instruction describing the target concept directly and a few instances exemplifying it. The system is required to incrementally learn a stream of tasks, so that the knowledge gained in the past can be used to address subsequent tasks. Apparently, this new problem tries to integrate the $\mathcal{S}_1$ and $\mathcal{S}_2$ into a single learning paradigm while decreasing the necessity of $\mathcal{S}_0$. More specifically, \taskname~ is expected to carry the properties listed in Table \ref{tab:properties}. 
\begin{table*}[ht]
 \setlength{\belowcaptionskip}{-10pt}
 \setlength{\abovecaptionskip}{5pt}
    \centering
    \small
    \begin{tabular}{l|l}
        Item & Explanation \\\hline
        Instruction-driven supervision & Each task is explained by an instruction and a couple of instances exemplifying it.\\
        Fixed model capacity & The system's structure and parameter size are constant regardless of its learning status. \\
        Knowledge maintenance & The system is not inclined to catastrophic forgetting. \\
        Forward transfer & The system uses knowledge acquired from upstream tasks to help solve downstream tasks.\\
        Backward transfer & The system uses knowledge acquired from downstream tasks to help solve upstream tasks.\\
        % Online learning & The model learns from a continuous data stream. No task boundaries The model learns without requiring neither clear task nor data boundaries.\\
        
    \end{tabular}
    \caption{Desiderata of \taskname, inspired by \cite{DBLPesialskaBC20}.}
    \label{tab:properties}
\end{table*}

Our data set is restructured from the \kashabidata\enspace \cite{DBLP08773}. \kashabidata\enspace is a benchmark that studies if a model can make appropriate use of natural language instructions to answer inputs accordingly. It comprises 61 tasks; each task is associated with a piece of instruction consisting of \texttt{Title}, \texttt{Definition}, \texttt{Caution}, \texttt{Prompt}, \texttt{Things to avoid}, \texttt{Examples}, etc. \kashabidata\enspace originally focuses on conventional supervised learning: give a bunch of tasks out of the 61 as the training tasks, and evaluate the remaining tasks  in a batch. In order to fit the formulation of \taskname, we reorganize the 61 tasks in \kashabidata:   a few tasks (e.g., size $k$) out of the 61 act as training tasks,  and the remaining $61-k$ tasks as an ordered list of new tasks. The learner is expected to first learn from the $k$ training tasks about how to use instructions to solve problems; then it evolves task by task along with the new task chain. 

% Specifically, we have two setups for \taskname. (i) \fewshottask: we randomly select a few tasks as training tasks to train the system at the initial stage, then let the system keep learning remaining tasks sequentially.  (ii) \zeroshottask: In this setup, no tasks are treated as training tasks particularly, therefore all the 61 tasks will act as new tasks to be learned in a continual learning paradigm. The  reasons that we have such separate setups are two-fold. Firstly, our goal is to study the model's ``incremental'' behavior  rather than batch learning capability over a fixed set of tasks. Secondly, if we use the majority of tasks as training tasks (just like what \kashabidata~ does), it turns out to be the conventional supervised fashion again: in order to check if a model can handle a new task from its constructions, we have to first prepare a large number of training tasks in this kind. The point of using instruction-driven supervision to decrease the demand of labeled examples vanishes.

Our system \modelname\enspace is based on BART \cite{DBLPLGGMLSZ20} with two proposed strategies aiming at making the best use of instructions. The first strategy, ``\negativetraining'', makes use of unfavorable clues, such as \texttt{Things to avoid}, from the instruction to promote the task understanding and forward-transfer. The second strategy, ``\historytraining'', revisits instructions of earlier tasks during continual learning  to alleviate the catastrophic forgetting issue in backward-transfer. We evaluate \modelname~on a wide range of transferring distances (from 1 to 40), which shows that \modelname~can generally help both forward-transfer and backward-transfer.\footnote{"Transferring distance"  refers to the task numbers between  the model at a new status and the model at an earlier status.}

Overall, this work has made three-fold contributions. First, \taskname\enspace is the first time to be formulated and studied in the NLP community. Second, we  propose  \modelname, a promising approach to \taskname. Third, we conduct intensive analyses, aiming to give a better understanding of this new challenge.

\begin{figure*}[t]
\centering
\includegraphics[width=0.98\linewidth]{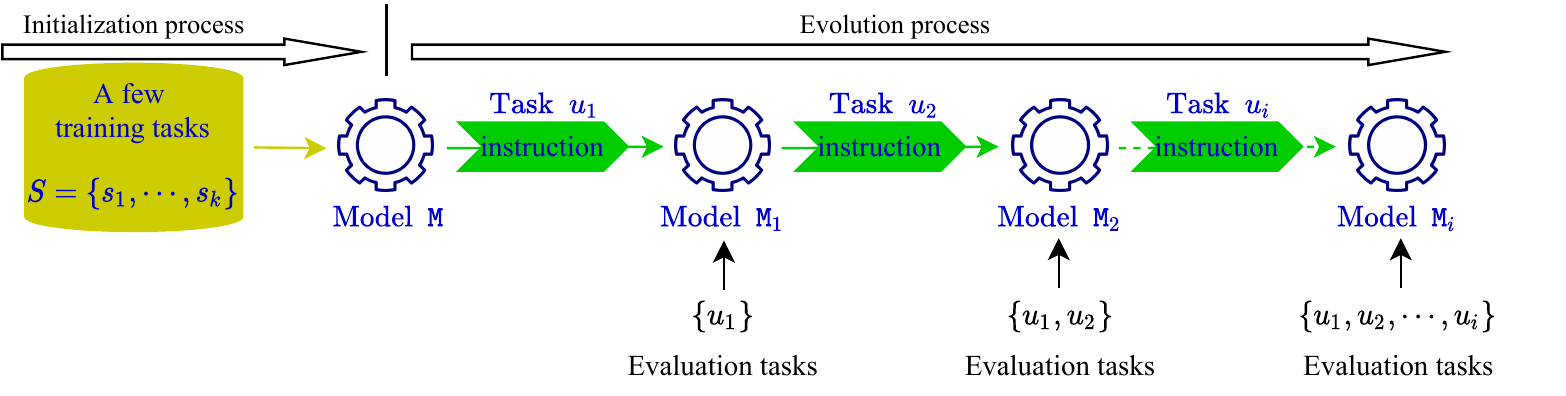}
\caption{The setup in \taskname. The whole learning process consists of two stages: \textit{initialization process} and \textit{evolution process}. A few training tasks $S=\{s_1, \cdots, s_k\}$ equipped with instructions and labeled examples are adopted to initialize the model \texttt{M}, then \texttt{M} incrementally learns from each unseen task $u_i$ by its instruction only. Once finishing the continual learning on the task $u_i$, the model \texttt{M}$_i$ is expected to be able to evaluate on all $\{u_1, u_2, \cdots, u_i\}$. } \label{fig:taskfigure}
\end{figure*}

\section{Related Work}
This section retrospects \textit{continual learning} and \textit{learning from task instructions},  two  machine learning paradigms that try to explore supervisions $\mathcal{S}_1$ and $\mathcal{S}_2$, respectively.

\paragraph{Continual learning.} Since the advent of continual learning\footnote{Continual learning in the literature is also referred to as: lifelong learning \cite{DBLPSilverM02}, incremental learning \cite{solomonoff1989system}, sequential learning \cite{mccloskey1989catastrophic}, and never-ending learning \cite{DBLPnBKSHM10}.} \cite{DBLPThrunM95}, this learning problem was mainly studied in computer vision or robotics domains, and most work concentrated on mitigating  catastrophic forgetting  \cite{mccloskey1989catastrophic,DBLPrraSMK18,DBLPfmanningerPBP20}. Continual learning can be summarized into three categories: class continual learning (\texttt{CCL}), domain continual learning (\texttt{DCL}), and task continual learning (\texttt{TCL}). 

\texttt{CCL} learns a sequence of classes (e.g., visual object categories, text labels, etc.)  to build one overall multi-label classifier for all the classes seen so far \cite{DBLPYanX021}. For example, \newcite{DBLPgXYGCW19} studied incrementally learning new relations for two entity mentions in an input sentence, and each relation has many labeled examples. 
% Their method tried to keep some examples of previous relations to ease the catastrophic forgetting. 
\newcite{DBLPXiaYFY21} proposed few-shot \texttt{CCL} in which multi-round of new text tags (e.g., intents or relations expressed in the input text) are encountered sequentially, and each new tag is only accompanied by a couple of examples.   

\texttt{DCL}  essentially studies the same task but in different domains. The system is expected to evolve along with learning from a stream of datasets of the same task and different data distributions. Typical work in NLP includes sentiment classification \cite{DBLPnM015,DBLPXiaJH17}, conversational agents \cite{DBLP09943}, text classification, and question answering \cite{DBLPumeRKY19}, etc.

\texttt{TCL} tries to learn distinct tasks sequentially.  Systems in  \cite{DBLPunHL20,DBLPunWZZ20}   incrementally learned among five disparate NLP tasks.
% (question answering, semantic parsing, sentiment analysis, semantic role labeling and goal-oriented dialogue).
\newcite{ntinually} further extended the size of the task stream (one benchmark has 26  tasks, the other covers 55) and studied  \texttt{TCL} in a few-shot scenario.  It is worth mentioning that all the listed work in \texttt{TCL} consistently transformed all tasks into question answering format (as pointed out in \cite{DBLP8730}, many NLP tasks can be formulated as
question answering),  thus  \texttt{TCL} in these literature was actually converted into \texttt{DCL}.

Similar with \cite{DBLPXiaYFY21, ntinually}, our work also focuses on low-resource continual learning; in contrast, our learning problem belongs to \texttt{TCL} while \textit{each task in our formulation is expressed by instructions instead of labeled examples}.

% \paragraph{Continual learning algorithms.} (i) ``rehearsal'' and ``pseudo-rehearsal''. both rehearsal and pseudo-rehearsal approaches imply some form of concurrent learning. (ii) ``structural regularization'': one seeks to prevent major changes in the weights that were important for previous tasks. Dedicating specific sub-parts of the network for each task is another way of reducing representational overlap.

% \newcite{DBLPunHL20} worked on \texttt{TCL}; all tasks are converted into QA and a single language model is used for the continual learning; before training on a new task, the language model first generate pseudo-examples for previous tasks; those pseudo-examples are mixed with the examples of the new task to train the language model. 

\paragraph{Learning from textual instructions.} This learning paradigm was first presented by \newcite{DBLPldwasserR14}. They investigated the challenges on Solitaire card game where an instruction is a short sentence such as ``\textit{you can move any top card to a free cell if it is empty}'', then this instruction is mapped into logical expression via semantic parsing so that an automated agent can understand and execute the instruction. 

More recent work tried to examine the ability of large-scale pretrained language models to follow natural language instructions of varying complexity. For example, \newcite{DBLP982} tested GPT-2 \cite{radford2019language} to understand instructions like ``\textit{listing nouns}'', ``\textit{output the $n$th word or char}'' and real-world MTurk instructions to annotate some popular datasets. They concluded that GPT-2 works poorly when the supervision comes from those instructions. A dominant instruction format nowadays is called ``prompt'' which mostly is a short piece of text describing the core concept of the task. Representative work includes \cite{radford2019language,DBLP1926,DBLPchickS21}, etc. (Please refer to the survey \cite{DBLP586} for more details.) 

While these prompt-based results are encouraging, such prompts are often too simplistic, whereas many real NLP problems cannot be effectively formulated as short prompts or a few positive examples. Motivated, \newcite{DBLP08773} collected more than 60 distinct NLP tasks with real-world MTurk instructions, and claimed that pretrained language models, such as BART  and GPT-3 \cite{DBLPBrownMRSKDNSSAA20}, benefit from instructions to generalize across tasks.

To our knowledge, the only work somehow resembling ours is \cite{DBLPRostamiIE20}, in which task descriptions were incorporated into lifelong learning for zero-shot transfer. We differ in three aspects: (i) they focused on robot controlling problems, (ii) their tasks are from a single domain, and (iii) in addition to the associated instruction, they assumed that each task has a large number of labeled examples.

\section{Problem formulation}

\subsection{\taskname}
A system in our \taskname~comprises two stages, as illustrated in Figure \ref{fig:taskfigure}. The first stage describes its starting status before learning the first new task; the second stage describes how it evolve continually with a sequence of instruction-equipped unseen tasks. To make it easier to understand, we first introduce the \textit{evolution process}, then the \textit{initialization process}. 

\paragraph{Evolution process.} \taskname~ tries to build a model $\texttt{M}$ that is able to deal with  unseen tasks ($U$)  appearing consecutively by understanding merely the instruction of each task. We denote the task sequence as $U$ = [$u_1$, $u_2$, $\cdots$, $u_i$, $\cdots$]. Each task $u_i$ has a piece of textual description $d_{u_i}$, and a set of evaluation instances $\{(x_{u_i}^j, y_{u_i}^j)\}_{j=1}^n$ where $y_{u_i}^j$ is the expected output of the input $x_{u_i}^j$. An  example $d_{u_i}$ will be shown in Section \ref{sec:data}. We denote the model \texttt{M}, having learned [$u_1$, $\cdots$, $u_i$], as \texttt{M}$_i$. For each task $u_i$,  \texttt{M}$_i$ is required to generate the output for $x_{u_i}^i$ based on the instruction in $d_{u_i}$.

\paragraph{Initialization process.} How to teach a system some basic knowledge to understand task instructions and learn continually? We prepare a few training tasks ($S$=[$s_1$, $s_2$, $\cdots$, $s_k$]) to equip the machine with the ability to  annotate the task instances given instructions. Each training task $s_i$ also has its instruction $d_{s_i}$ and $n$ labeled examples \{($x_{s_i}^j$, $y_{s_i}^j$)\}$_{j=1}^n$. Note that here we want to control $k$ to be  small; otherwise, if \taskname~ requires a large number of training tasks at the initialization stage, there is no point anymore to make use of instructions to alleviate the burden of data annotation.

\subsection{Evaluation protocol}\label{sec:evaluationprotocol}

\paragraph{Forward-transfer evaluation.}

% \begin{algorithm}[t] 
% \caption{Forward-transfer metric calculation}
% \label{alg:forwardmetric}
% \begin{algorithmic}[1]
% \Require{The model \texttt{M}, all unseen tasks in $U$, two hyperparameter $m$ and $i$} 
% \Ensure{$\overrightarrow{g}_i$}
% % \Statex
% \For{task $t$ in $U$}                    
%     % \State {put $t$ at position $i+1$};
%     \For{$j<m$ times}                    
%     \State {sample [$u_1$, $\cdots$, $u_{i}$] from $U-\{t\}$;}
%     \State {\texttt{M}$_i$ = \texttt{M} evolves over [$u_1$, $\cdots$, $u_{i}$, $t$];}
%     \State {$\overrightarrow{p}_{i,t}^j$=\texttt{M}$_i$($t$);}
%     \EndFor
%     \State {$\overrightarrow{p}_{i,t}=\frac{1}{m}\sum_{j=1}^m{\overrightarrow{p}_{i,t}^j}$};
% \EndFor
% \State {$\overrightarrow{g}_i=\frac{1}{|U|}\sum_{t\in U}{(\overrightarrow{p}_{i,t}-\overrightarrow{p}_{0,t})}$}
% \end{algorithmic}
% \end{algorithm}

\begin{algorithm}[t] 
\caption{Forward-transfer metric calculation}
\label{alg:forwardmetric}
\begin{algorithmic}[1]
\Require{The model \texttt{M}, all unseen tasks in $U$, two hyperparameter $m$ and $i$} 
\Ensure{$\overrightarrow{g}_i$}
% \Statex
\For{task $t$ in $U$}                    
    \For{$j<m$ times}               
    \State {$k$ = random.randint(1, |U|-$i$)};
    \State {sample [$u_1$, $\cdots$, $u_{k-1}$, $u_{k}$, $\cdots$, $u_{k+i-1}$] from $U$-\{t\};}
    \State {\texttt{M}$_k$ = \texttt{M} evolves over [$u_1$, $\cdots$, $u_{k-1}$, $t$];}
    \State {\texttt{M}$_{k+i}$ = \texttt{M} evolves over [$u_1$, $\cdots$, $u_{k+i-1}$, $t$];}
    \State {$\overrightarrow{g}_{i,t}^j$=\texttt{M}$_{k+i}$($t$) - \texttt{M}$_{k}$($t$);}
    \EndFor
    \State {$\overrightarrow{g}_{i,t}=\frac{1}{m}\sum_{j=1}^m{\overrightarrow{g}_{i,t}^j}$};
\EndFor
\State {$\overrightarrow{g}_i=\frac{1}{|U|}\sum_{t\in U}{\overrightarrow{g}_{i,t}}$}
\end{algorithmic}
\end{algorithm}

\begin{algorithm}[t] 
\caption{Backward-transfer metric calculation}
\label{alg:backwardmetric}
\begin{algorithmic}[1]
\Require{The model \texttt{M}, all unseen tasks in $U$, two hyperparameter $m$ and $i$} 
\Ensure{$\overleftarrow{g}_i$}
% \Statex
\For{task $t$ in $U$}                    
    \For{$j<m$ times}               
    \State {$k$ = random.randint(1, |U|-$i$)};
    \State {sample [$u_1$, $\cdots$, $u_{k-1}$, $t$, $u_{k+1}$, $\cdots$, $u_{k+i}$] from $U$;}
    \State {\texttt{M}$_k$ = \texttt{M} evolves over [$u_1$, $\cdots$, $u_{k-1}$, $t$];}
    \State {\texttt{M}$_{k+i}$ = \texttt{M} evolves over [$u_1$, $\cdots$, $u_{k+i}$];}
    \State {$\overleftarrow{g}_{i,t}^j$=\texttt{M}$_{k+i}$($t$) - \texttt{M}$_{k}$($t$);}
    \EndFor
    \State {$\overleftarrow{g}_{i,t}=\frac{1}{m}\sum_{j=1}^m{\overleftarrow{g}_{i,t}^j}$};
\EndFor
\State {$\overleftarrow{g}_i=\frac{1}{|U|}\sum_{t\in U}{\overleftarrow{g}_{i,t}}$}
\end{algorithmic}
\end{algorithm}

For this metric, we attempt to quantify the effectiveness of learning more prior tasks before solving a target task. Intuitively, more prior tasks, better downstream performance.  We define metric  $\overrightarrow{g}_i$ (hereafter,  ``$\rightarrow$'' refers to forward-transfer and ``$\leftarrow$'' refers to backward-transfer): the \textit{average gained performance} over all new tasks in $U$ when each of them is learned after $k+i-1$ previous tasks, compared with learning them merely after $k-1$ tasks ($i$ is transferring distance). As Algorithm \ref{alg:forwardmetric} shows, computing $\overrightarrow{g}_i$ needs two loops. First, iterate on all tasks in $U$, select one task $t$ as (i) the $k$th task and randomly sample its upstream tasks [$u_1$, $\cdots$, $u_{k-1}$] from remaining tasks in $U$, getting one online learning score $\texttt{M}_k(t)$, or as (ii) the $(k+i)$th task for another online learning score $\texttt{M}_{k+i}(t)$. $\texttt{M}_{k+i}(t)$ - $\texttt{M}_k(t)$ is one instance of the forward-transfer score, which indicates how much improvement the extra upstream tasks of size $i$ bring to the target task $t$. For this particular task $t$, repeat its upstream tasks $m$ times and calculate the average as a final score of $t$, denoted as $\overrightarrow{g}_{i,t}$. Second, the same procedure is applied to all tasks in $U$ and finally average $\overrightarrow{g}_{i,t}$ over all $t$ to get the $\overrightarrow{g}_i$ value.

$\overrightarrow{g}_i$ always measures the expected performance gain our system can get when it has continually leaned $i$ more tasks.  For forward-transfer, we expect $\overrightarrow{g}_i$ is positive and increases when $i$ gets larger.

\paragraph{Backward-transfer evaluation.} In contrast to the forward-transfer evaluation, we define $\overleftarrow{g}_i$ as the backward-transfer metric, which tells how much better our system can handle a task learned $i$ steps ago, compared with the performance on the same the task last time. As Algorithm \ref{alg:backwardmetric} describes,  two loops to calculate $\overleftarrow{g}_i$. Firstly, for a given task $t$ from $U$, put $t$ in a random position $k$ in the task chain, followed by $i$ other tasks. Subtract its performance when the model \texttt{M} learned it the first time (i.e., \texttt{M}$_k(t)$) by its performance when the model finished learning all the $k+i$ tasks in the chain (i.e., \texttt{M}$_{k+i}(t)$). This operation generates a score given this chain; repeat this process $m$ times to get an average gain $\overleftarrow{g}_{i,t}$ for the task $t$. Secondly, average the  $\overleftarrow{g}_{i,t}$ over all $t$ to get the $\overleftarrow{g}_i$ value.

If a system can always make use of downstream tasks to help upstream tasks, $\overleftarrow{g}_i$ should be positive; otherwise, $\overleftarrow{g}_i$ will be negative due to catastrophic forgetting.

\subsection{Data} \label{sec:data}

There are no NLP datasets for \taskname~particularly. This work is based on \kashabidata\enspace \cite{DBLP08773} after data reorganization. Next, we first introduce \kashabidata, then describe our revised version specific to our problem.

\begin{figure}[t]
 \setlength{\belowcaptionskip}{-10pt}
 \setlength{\abovecaptionskip}{5pt}
\centering
\includegraphics[width=0.9\linewidth]{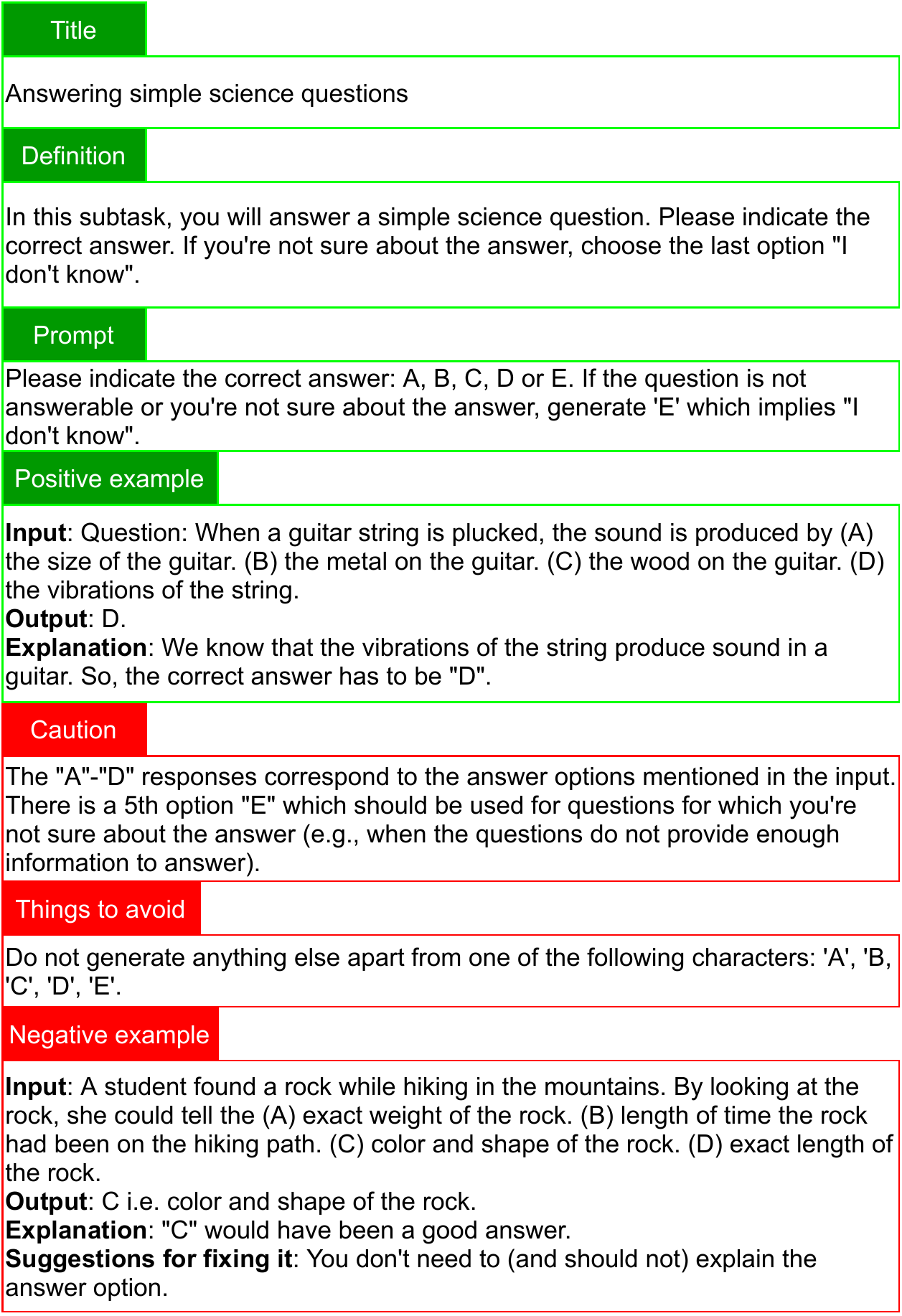}
\caption{An instruction example for the task ``answering science questions (misc)'' in \kashabidata~\cite{DBLP08773}. \textcolor{green}{Green} parts present favorable clues while \textcolor{red}{red} parts express unfavorable predictions.} \label{fig:instruction}
\end{figure}

\kashabidata~was constructed in the following pipeline: \newcite{DBLP08773} first collected some popular NLP benchmarks (e.g., CosmosQA \cite{DBLPangBBC19}, Quoref \cite{DBLPgiLMSG19}, Winogrande \cite{DBLPguchiBBC20}, etc.) with their crowdsourcing instructions through engaging with their authors. Since all the crowdsourcing instructions include multiple steps to guide annotators to gather task instances, they further broke raw crowdsourcing instructions down into their individual steps, generating a larger number of  subtasks that are  minimal and standalone. At last, a total of 61 tasks are obtained, covering six categories: 13 question generation tasks (QG), 16 answer generation tasks (AG), 12 classification tasks (CF), 8 incorrect answer generation tasks (IAG), 10 minimal modification tasks (MM) and 2 verification tasks (VF). An instruction example is presented in Figure \ref{fig:instruction}.

\paragraph{Our data split.} For training tasks $S$, we randomly select $k$ tasks from the 61 tasks. All training tasks in $S$ have instructions and keep their labeled example set. The remaining 61-$k$ tasks are treated as unseen task set $U$. Each task in $U$ has only instruction; the labeled example set is used for evaluation rather than model training.

It is noteworthy that task order in continual learning should influence the final performance. We do not attempt to release a fixed split of $S$ and $U$.  In experiments, we will randomly generate them multiple times to form different task chains and report the average performance.  

\section{Our method \modelname}

Most prior studies about continual learning focused on backward-transfer  \cite{DBLPrraSMK18,DBLPumeRKY19} while paying less attention to the forward-transfer performance. 
% As our evaluation protocol (Section \ref{sec:evaluationprotocol}) indicates, both forward-transfer and backward-transfer performances are of our interest in this \taskname~ problem. 
Next, we introduce our approach to promoting both of them.

\textit{The big story of our strategies lies in better understanding  of the textual instruction of $u_i$}. Two concrete strategies as follows.

\paragraph{\negativetraining:} \textit{to distinguish favorable and unfavorable clues in instructions}.
Unfavorable clues, such as the red items in Figure \ref{fig:instruction}, are essential for humans to make decisions while not being successfully leveraged by machine learning. For example, \newcite{DBLP08773} found discarding negative examples can even improve the performance. We believe this indicates the approach failed to learn from negative examples rather than those examples being truly useless. Then, how can we make machines  extract effective supervision from negative samples?

First, we introduce a method  that was tried but did not work well -- \textit{minimizing the probability of generating negative output}. Maximizing the probabilities of gold output is widely used in text generation. It sounds intuitive to minimize that for unwanted output, such as \cite{DBLPHeG20}. We tried a joint training on maximizing positive and minimizing negative examples, which is even worse than maximizing the positive alone. Since many ``negative'' outputs contain tokens that exist in the gold answers,  we suspect that minimizing their probabilities will let the model have more difficulty  decoding the correct output. 

% (ii) \textit{Build a negative-output generator given available negative examples in instructions}. We planed to use this type of negative output to replace the instruction-provided negative examples for pretraining. However, due to the tiny size of negative examples in instructions (most tasks have at most 2 negative examples, a couple of them have zero), the learned negative-output  generator yields outputs that are over unreasonable.

After further study of those negative examples and their explanations, we decide to treat those negative examples as positive and move the negative learning phase as pretraining, i.e.,  pretrain on negative examples first, then finetune on positive examples. The inspiration comes from the fact that negative examples, despite the tag ``negative'', can still provide useful information about the expected output. Take a look at the negative example in Figure \ref{fig:instruction}, its output ``C i.e., color and shape of the rock'' is discouraged just because it does not follow some rules of automatic evaluation rather than  it is really wrong. Apparently, as a first step, optimizing the system to generate the so-called ``negative output'' is still better than any general-purpose pretrained BART.

For each unseen task in $U$, we directly adopt its negative examples if available. For the $k$ training tasks in $S$, positive instances (including positive examples in instructions and those labeled task instances) are much more than the negative examples, we use the pretrained model on $S$ to do prediction on all inputs of $S$, if the output is not equal to the gold output, we treat this (input, predicted output) as a negative example. It means we have a loose definition of what ``negative output'' is: it is negative once it is not equal to the ground truth. Since the pretrained model on $S$ can already guarantee generation quality, those generated negative outputs are mostly related with the gold outputs (measured by ROUGE metrics).\footnote{We also tried to \textit{build a negative-output generator given available negative examples in instructions}. This type of negative output was planed for pretraining in both $S$ and $U$. However, due to the tiny size of negative examples in instructions (most tasks have at most 2 negative examples, a couple of them have zero), the learned negative-output  generator yields outputs that are over unreasonable.}

\paragraph{\historytraining:} \textit{revisit instructions of previous tasks}. To mitigate catastrophic forgetting, many prior works  about continual learning tried to store a couple of labeled examples of upstreaming tasks to replay. In our \taskname~formulation, each new task is described merely by the instruction. Instead of storing some examples of previous tasks,  we keep their instructions. When learning the $i$th task in $U$, our model will first learn all the instructions of prior $i-2$ tasks in a batch with a lower learning rate. Revisiting precedent instructions is cost-effective  since each instruction is as short as a couple of conventionally annotated examples but with much more supervision.

Overall, our two strategies work jointly to enhance the forward-transfer    and the backward-transfer performance. Our system \modelname\enspace is based on BART, treating all tasks as a text-to-text problem. The full input format of encoder: \textcolor{purple}{[\texttt{Input}] input string} \textcolor{orange} {[\texttt{Title}] title string [\texttt{Prompt}] prompt string [\texttt{Definition}] definition string [\texttt{Avoid}] things to avoid string [\texttt{Caution}] caution string} \textcolor{blue}{[\texttt{POS1}] [\texttt{Input}] input string [\texttt{Output}] output string [\texttt{Explanation}] explanation string $\cdots$ [\texttt{POS$n$}] [\texttt{Input}] input string [\texttt{Output}] output string [\texttt{Explanation}] explanation string}. Note that we put the input at the beginning of this encoder's input template to prevent from being discarded due to long text truncation. When pretrain on training tasks $S$, the full input pattern is used; when continually learn on $U$, since the \textcolor{purple}{input} at the beginning comes from positive or negative examples of the instruction, we do not include the positive examples in the input template (i.e., the blue part is dropped).

Given $S$ and $U$, the whole learning pipeline in \modelname~is: (i) pretrain on $S$ to get model \texttt{M}$^*$; (ii) use \texttt{M}$^*$ to make predictions on $S$ to collect negative example set $S^{-}$; (iii) pretrain on $S^{-}$ and finetune on $S$ to get boosted model \texttt{M} which is the starting model status for continual learning on $U$; (iv) for the $i$th unseen task $u_i$ in $U$, tune \texttt{M} on instructions of all  earlier tasks [$u_1$, $\cdots$, $u_{i-2}$] in a batch; (v) tune on negative examples of $u_i$, if available; (vi) tune on positive examples of $u_i$.

\section{Experiments}
\begin{table*}[ht]
\setlength{\tabcolsep}{2.7pt}
    \centering
    % \small
  \begin{tabular}%
  {>{\raggedright\arraybackslash}p{0.28\linewidth}||%
   >{\raggedleft\arraybackslash}p{0.055\linewidth}%
   >{\raggedleft\arraybackslash}p{0.055\linewidth}%
   >{\raggedleft\arraybackslash}p{0.055\linewidth}%
   >{\raggedleft\arraybackslash}p{0.055\linewidth}%
   >{\raggedleft\arraybackslash}p{0.055\linewidth}||%
   >{\raggedleft\arraybackslash}p{0.055\linewidth}%
   >{\raggedleft\arraybackslash}p{0.055\linewidth}%
   >{\raggedleft\arraybackslash}p{0.055\linewidth}%
   >{\raggedleft\arraybackslash}p{0.055\linewidth}%
   >{\raggedleft\arraybackslash}p{0.055\linewidth}%
  }
    \multirow{2}{*}{Method} & \multicolumn{5}{c||}{forward-transfer} & \multicolumn{5}{c}{backward-transfer}\\
     & $\overrightarrow{g}_1$ & $\overrightarrow{g}_{10}$ & $\overrightarrow{g}_{20}$ & $\overrightarrow{g}_{30}$ & $\overrightarrow{g}_{40}$ & $\overleftarrow{g}_1$ & $\overleftarrow{g}_{10}$ & $\overleftarrow{g}_{20}$ & $\overleftarrow{g}_{30}$ & $\overleftarrow{g}_{40}$\\\hline
    
    Seq-finetune & 1.44 \scriptsize{$\pm$7.15} & 3.28  \scriptsize{$\pm$19.46} & -3.74  \scriptsize{$\pm$8.73} & 2.9  \scriptsize{$\pm$16.42} & -0.36  \scriptsize{$\pm$17.23}  &1.57 \scriptsize{$\pm$3.28} & 0.04  \scriptsize{$\pm$12.46} & -0.19  \scriptsize{$\pm$21.75} & -6.48  \scriptsize{$\pm$19.17} & -9.46  \scriptsize{$\pm$19.57} \\
    
    LAMOL \cite{DBLPunHL20} &-1.34  \scriptsize{$\pm$4.46}& 1.41  \scriptsize{$\pm$13.55}  & 3.31  \scriptsize{$\pm$14.32} & -5.40  \scriptsize{$\pm$20.44} & -0.03  \scriptsize{$\pm$12.68} &2.67  \scriptsize{$\pm$12.52}& 2.21  \scriptsize{$\pm$7.98}  & 9.42  \scriptsize{$\pm$12.88} & 6.33  \scriptsize{$\pm$20.13} & 7.21  \scriptsize{$\pm$14.81} \\\hline
    
    Our \modelname & \textbf{2.16}  \scriptsize{$\pm$6.46}& \textbf{5.06}  \scriptsize{$\pm$20.87}  & 2.29  \scriptsize{$\pm$18.03} & \textbf{4.07}  \scriptsize{$\pm$7.95} & \textbf{4.39}  \scriptsize{$\pm$14.56} &1.44  \scriptsize{$\pm$9.28}& \textbf{5.21}  \scriptsize{$\pm$18.20}  & 7.33  \scriptsize{$\pm$13.48} & \textbf{14.99}  \scriptsize{$\pm$20.21} & \textbf{12.31}  \scriptsize{$\pm$16.53}\\
    
    \enspace\enspace w/o \negativetraining & -2.89  \scriptsize{$\pm$13.12} &1.06  \scriptsize{$\pm$17.21}  & 1.33  \scriptsize{$\pm$13.09} & 2.21  \scriptsize{$\pm$14.42} & 1.78  \scriptsize{$\pm$17.90} &2.21  \scriptsize{$\pm$12.23}& 3.37  \scriptsize{$\pm$13.23}  & \textbf{11.44}  \scriptsize{$\pm$11.03} & 10.36  \scriptsize{$\pm$21.34} & 8.94  \scriptsize{$\pm$19.41}\\
    
    \enspace\enspace w/o \historytraining &1.88  \scriptsize{$\pm$17.73} &3.32  \scriptsize{$\pm$12.76}  & \textbf{4.41}  \scriptsize{$\pm$20.24} & 3.22  \scriptsize{$\pm$16.66} & 2.97  \scriptsize{$\pm$14.93} &\textbf{4.74} \scriptsize{$\pm$16.54} & -2.78 \scriptsize{$\pm$19.38} & -0.83  \scriptsize{$\pm$12.93} & 1.35 \scriptsize{$\pm$15.95} & 3.49  \scriptsize{$\pm$14.05} \\\hline
    
    Multi-task (upperbound) &\multicolumn{10}{c}{\enspace\enspace\enspace 7.98\scriptsize{$\pm$20.47}}  \\
        
    \end{tabular}
    \caption{The main  results of \taskname.}
    \label{tab:mainresult}
\end{table*}

\begin{table*}[ht]
    \centering
    % \small
    \begin{tabular}{ll||rrrrrr|c}
    \multicolumn{2}{l||}{Method} & QG & AG & CF & IAG & MM & VF & mean \\\hline
    \multirow{2}{*}{\cite{DBLP08773}} & paper report &52.xx & 30.xx & 50.xx &25.xx & 47.xx & 8.xx & 35.33 \\
        & reimplement & 53.55 &  17.45 & 63.79  & 11.06 &  82.86 & 7.40  &  39.35 \\\hline
        % pretrain on neg & 55.78 & 24.74 & 61.20 & 12.60 & 83.77 & 8.23 & 41.05 \\
    \multirow{2}{*}{Seq-finetune}&forward& 49.61  & 21.46  & 48.74  & 9.70 &57.31  &7.61  & 32.40 \\
    &backward &47.09  &21.17  &7.45  & 9.61 & 88.84 & \textbf{14.98} & 31.52\\\hline

    \multirow{2}{*}{LAMOL}& forward & 52.23 & 20.45 & 67.74 &8.81 & 82.29& 8.83 & 40.05\\
    & backward & 52.14 & 22.76 & 7.98& 8.33 & 88.45& 9.91&31.59 \\\hline
            \multirow{3}{*}{\modelname} &w/o CL & 51.07 & 23.40 & 70.68 & \textbf{11.43} & 88.13 & 6.22 & 41.82\\
    &forward &51.30  &24.89  &\textbf{70.96} & 9.36& \textbf{90.41}  &10.70  & \textbf{42.93}\\
    & backward &53.04  & \textbf{24.93} &7.51  & 8.56 &88.09  & 13.86 & 32.66\\

    \end{tabular}
    \caption{The results on standard split of \kashabidata. We use ``52.xx'' just because the original paper by \newcite{DBLP08773} did not report the ``xx'' numbers.}
    \label{tab:benchmarkresult}
\end{table*}

\paragraph{Setup.}  We use the pretrained BART-base model released by Huggingface. Hyperparameters: $m=10$ in Algorithms \ref{alg:forwardmetric}-\ref{alg:backwardmetric};  $k=5$ for the task set $S$;  max input length 1024 tokens, learning rate 5e-5, 3 epochs as suggested by \cite{DBLP08773} for most phases of training (except for 5e-6 and one epoch for \historytraining); batch size 5 for training on $S$ and 2 for continual learning on $U$. All unseen tasks $U$ randomly select 1k labeled examples for performance evaluation. Note that the official evaluation metric for \kashabidata\enspace is ROUGE-L \cite{lin2004rouge}. According to the definitions of our evaluation metrics, $\overrightarrow{g}_i$ and $\overleftarrow{g}_i$ numbers are  the same meaning as ROUGE-L.

\paragraph{Baselines.} There are no prior systems that can fit the formulation of \taskname~exactly. In addition,  as the \taskname~properties in Table \ref{tab:properties} indicate, ideally, \taskname~prefers a fixed model capacity. Therefore, we do not compare with  systems that incorporate extra memory modules or adaptors, such as \cite{DBLPumeRKY19,ntinually,DBLeXL21}. The following systems are considered:

\textbullet\enspace \textit{Seq-finetune}: first pretrain a BART on $S$, then fine-tune it on   $U$  sequentially. It does not pay special attention to catastrophic forgetting. 

\textbullet\enspace \textit{Multi-task}: first pretrain a BART on $S$, then train on instructions of all tasks in $U$ simultaneously.  It,  acting as the upperbound of continual learning, does not distinguish between forward-transfer and backward-transfer.

% \textbullet\enspace \textit{OEWC} \cite{DBLPz0LGTPH18}: Elastic Weight Consolidation (EWC, \cite{DBLPrkpatrickPRVD16}) regularizes the change of important model parameters during training. This online EWC (OEWC) keeps two components (a knowledge base and an active column), and optimize them  in two distinct, alternating phases. 
% In the \textit{progress} phase,  a new task is encountered, and only the active column gets optimised. When the progress phase is done, the active column is distilled into the knowledge base, thus forming the \textit{compress} phase.

\textbullet\enspace \textit{LAMOL} \cite{DBLPunHL20}: A state-of-the-art system that uses pretrained language models for task continual learning. 
All tasks are converted into QA and a single language model is used for the continual learning; before training on a new task, the language model first generates pseudo-examples for previous tasks; those pseudo-examples are mixed with the examples of the new task to train the language model. 
The original language model in LAMOL is a smallest pretrained GPT-2, we replace it with BART for a fair comparison. 

% \begin{figure*}[t]
%  \setlength{\belowcaptionskip}{-10pt}
%  \setlength{\abovecaptionskip}{5pt}
% \centering
% \subfigure[Forward-transfer] 
% { \label{fig:forward}
% \includegraphics[width=0.44\linewidth]{acl2022TACLNLItrainingSizeforward.pdf}
% }
% \subfigure[Backward-transfer] 
% { \label{fig:backward}
% \includegraphics[width=0.44\linewidth]{acl2022TACLNLItrainingSizebackward.pdf}
% }
% \caption{Influence of training task size (i.e., $k$ value in $S$=[$s_1$, $s_2$, $\cdots$, $s_k$]) on forward-transfer and backward-transfer. $k\in\{1,5,10,20\}$. Please note that $k=5$ is what we used to report Table \ref{tab:mainresult}.}
% \label{fig:influenceoftrainingsize}
% \end{figure*}

\begin{figure}[t]
\centering
\includegraphics[width=0.9\linewidth]{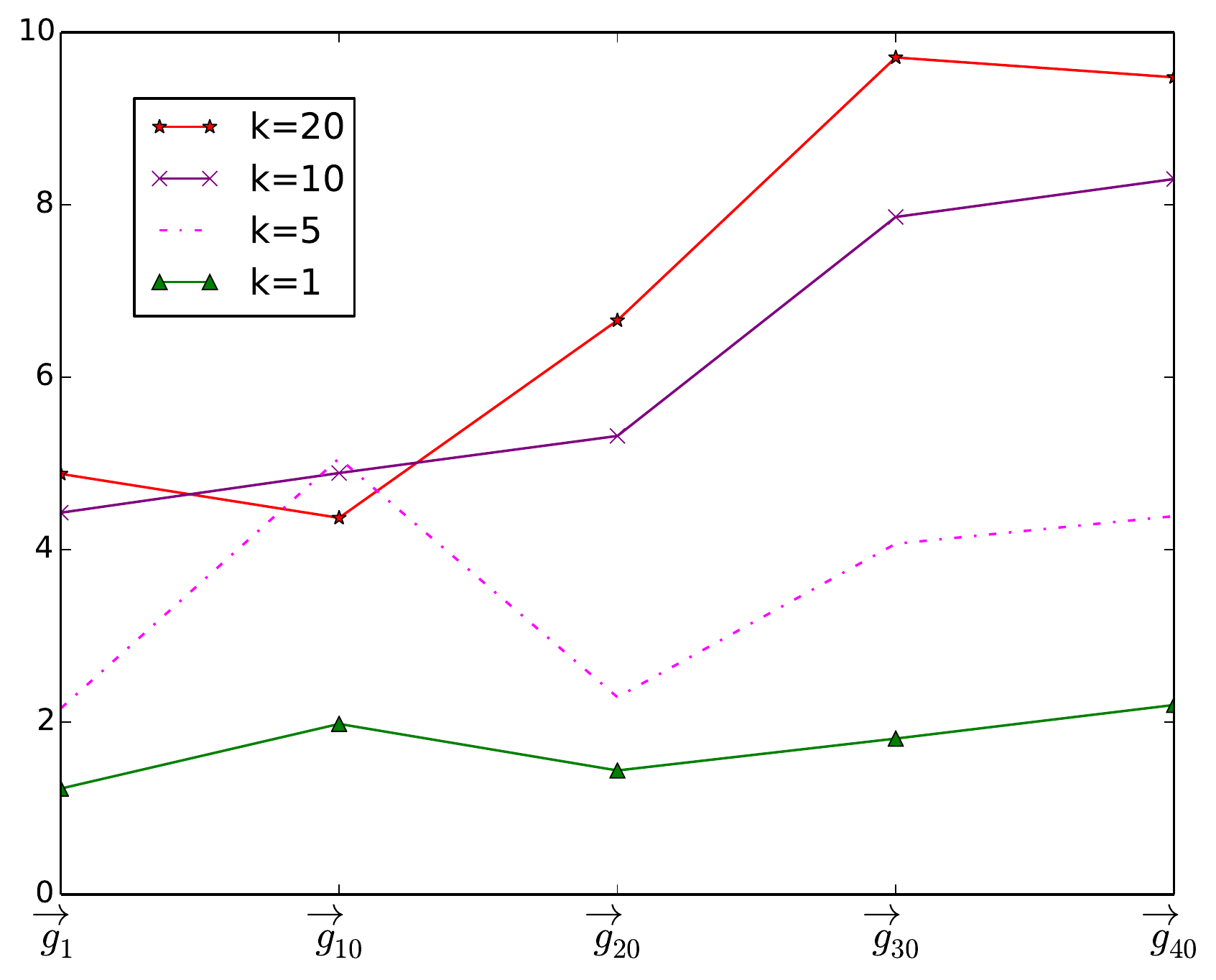}
\caption{Influence of training task size (i.e., $k$ value in $S$=[$s_1$, $s_2$, $\cdots$, $s_k$]) on forward-transfer. $k\in\{1,5,10,20\}$. Please note that $k=5$ is what we used to report Table \ref{tab:mainresult}.}
\label{fig:influenceoftrainingsize}
\end{figure}

\begin{figure*}[t]
 \setlength{\belowcaptionskip}{-10pt}
 \setlength{\abovecaptionskip}{5pt}
\centering
\subfigure[Forward-transfer] 
{ \label{fig:categoryforward}
\includegraphics[width=0.44\linewidth]{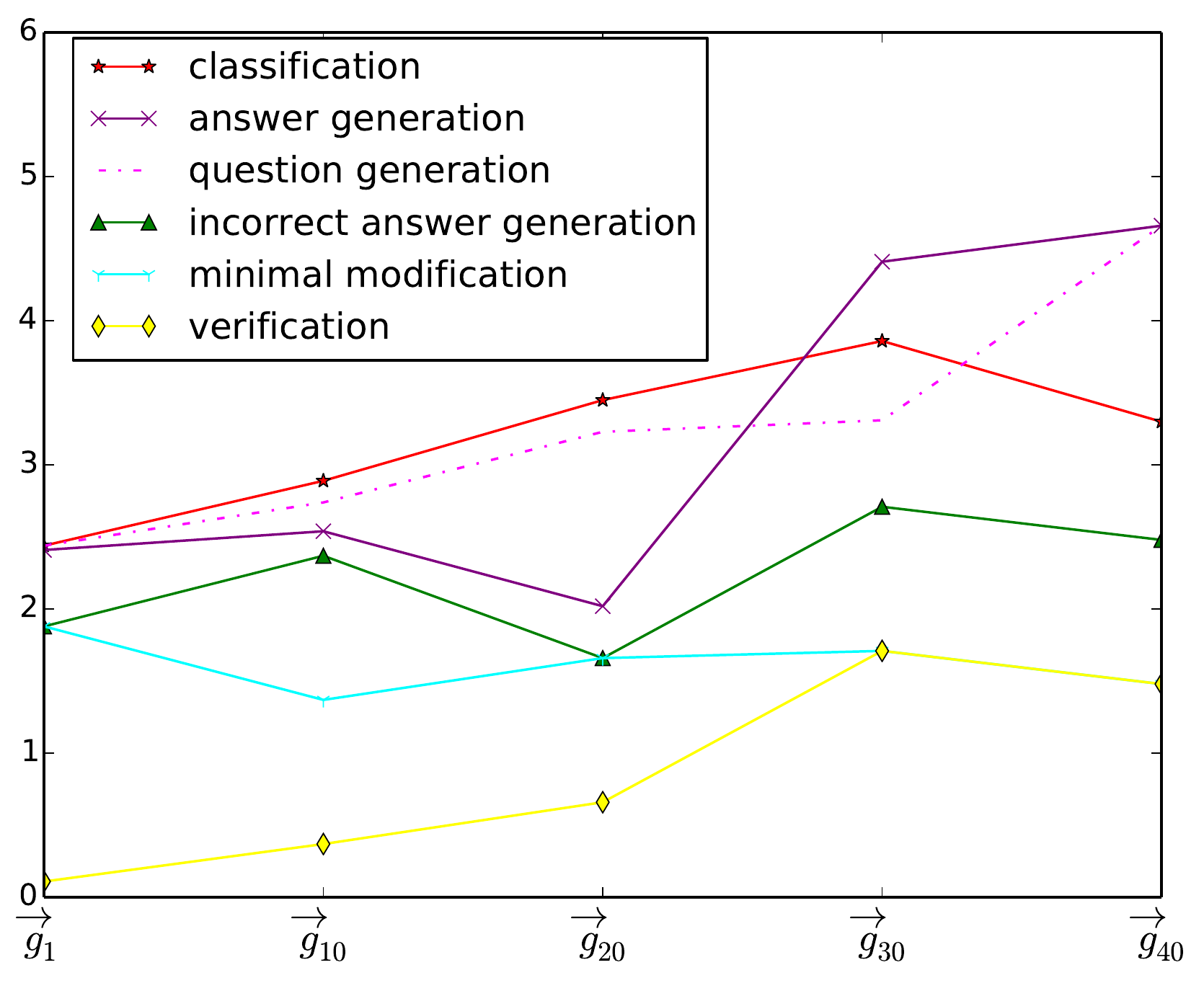}
}
\subfigure[Backward-transfer] 
{ \label{fig:categorybackward}
\includegraphics[width=0.44\linewidth]{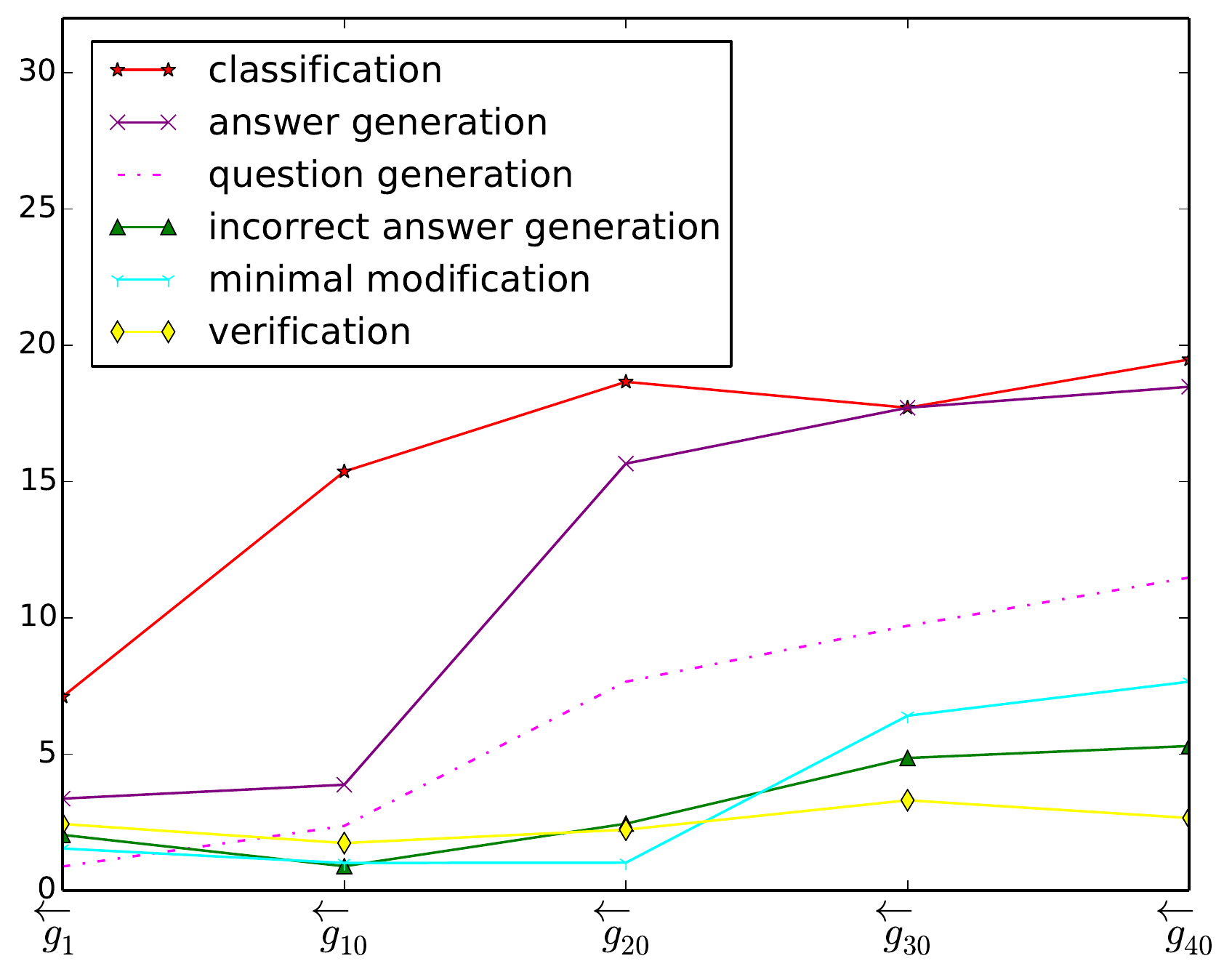}
}
\caption{Different transferabilities on tasks of six categories.}
\label{fig:influenceoftaskcategory}
\end{figure*}

% \begin{figure*}[t]
%  \setlength{\belowcaptionskip}{-10pt}
%  \setlength{\abovecaptionskip}{5pt}
% \centering
% \subfigure[Forward-transfer] 
% { \label{fig:itemforward}
% \includegraphics[width=0.44\linewidth]{acl2022TACLNLIinstructionitemforward.pdf}
% }
% \subfigure[Backward-transfer] 
% { \label{fig:itembackward}
% \includegraphics[width=0.44\linewidth]{acl2022TACLNLIinstructionitembackward.pdf}
% }
% \caption{Contributions of each constituent item in instructions.}
% \label{fig:influenceofitem}
% \end{figure*}

\subsection{Results}
Table \ref{tab:mainresult} shows the comparison between our system \modelname~and those baselines. We have three threads of observations.

Firstly, our system \modelname~consistently outperforms all baselines   for both forward-transfer and backward-transfer evaluations. For forward-transfer, all systems cannot beat the multi-task learning, but in backward-transfer, \modelname~even outperforms the multi-task competitor; this is because multi-task learning, though widely treated as upperbound for continual learning, only trained on all $U$ tasks for 3 epochs. Our method, equipped with \historytraining, actually learns many times of earlier $U$ tasks during the continual learning. Despite a few exceptions, generally speaking, both the forward and backward transfer performance increase when the transferring distance increases from 1 to 40. 

Secondly, the ablation study verifies the effectiveness of our two strategies. \negativetraining~plays the leading role in forward-transfer while doing a moderate favor to the backward-transfer. A totally opposite phenomenon is noticed for \historytraining: it clearly contributes to the backward-transfer evaluation while influencing the forward-transfer to some extent. 

Thirdly, the standard deviations are mostly large; this should be due to the fact that the 61 tasks in \kashabidata~ contains 6 distinct categories; each category benefits from the model generalization by different degrees. 

To further figure out the exact performance of our system on different task categories, we report on the standard split of \kashabidata~ as \newcite{DBLP08773} did: they have a fixed set of 12 tasks for testing (2 for each category), and all remaining tasks as training data. Since their 12 test tasks have no order, for each of the test category, we put it as the sixth  (resp. first) task in the chain for forward-transfer (resp. backward-transfer). Once the position of the test category is fixed, we randomly order the remaining five categories in the sequence for 10 times and report the average performance. Thus, each test category will have two numbers for every continual learning approach: one for forward-transfer, the other for backward-transfer. In addition, we also report our system \modelname~ without continual learning (w/o CL), i.e., using the system pretrained on 49 tasks in $S$ to predict. 

Table \ref{tab:benchmarkresult} lists the results of all continual learning systems on \kashabidata. We notice that (i) the results of different task  categories vary a lot. For example, minimal modification tasks (MM) easily get ROUGE-L score above 80, but it is pretty challenging to obtain  ROUGE-L score over 10 for Verification (VF); (ii) Classification tasks (CF) seem suffering from backward-transfer. We suspect CF is too sensitive to classification-specific supervision, such as label spaces; the continual learning on many subsequent tasks of different categories will mislead the model in solving CF. This is further supported by looking at the results of three systems: \modelname~ w/o CL, \cite{DBLP08773} and \modelname~forward-transfer. The first two systems start predicting on $U$ once finish the training on $S$.  Note that CF in $U$ has 10 CF tasks in $S$; it means the first two systems, although they did not learn the CF in $U$, still obtained enough supervision for this category from $S$. That's why all three systems get high performance on CF. Once they get tuned on more different categories, the supervision disappears increasingly.

\subsection{Analysis}

In addition to the  results in Tables \ref{tab:mainresult}-\ref{tab:benchmarkresult}, we are further interested in the following two questions.

\paragraph{$\mathcal{Q}_1$: how many training tasks does a system need to learn from instructions?} Recall that apart from the $U$ in the evolution process, we use $k$ tasks ($S$=[$s_1$, $s_2$, $\cdots$, $s_k$]) to initialize the model. $S$ can have maximal 20 tasks (due to the limited size of \kashabidata) and our system only used 5 out of them. Here, we further explore the model's behavior when $k$ varies.

Figure \ref{fig:influenceoftrainingsize}  depicts the influence of $k$ on forward-transfer. For forward-transfer, larger $k$ values (i.e., more training tasks to initialize the model) consistently improve performance. We think that more training tasks tend to teach the model better at understanding the task instructions, which can further improve the model's transferability when it learns  $i$ more tasks to report $\overrightarrow{g}_i$ on a downstream task $u_i$. We notice that \kashabidata~v2 \footnote{\url{https://github.com/allenai/natural-instructions}} has over 1.7k tasks. We leave it as future work to further explore the potential of increasing training tasks.

\paragraph{$\mathcal{Q}_2$: how do tasks of different categories in $U$ benefit?} In Section \ref{sec:data}, we mentioned that all tasks can be organized into  six categories.
% : question generation, answer generation, classification, incorrect answer generation, minimal modification and verification. 
We check their separate performances here. Note that both Algorithms \ref{alg:forwardmetric}-\ref{alg:backwardmetric} obtain the final score by averaging over all tasks in $U$, here we average those tasks that belong to the same category to get category-wise forward-transfer and backward-transfer performances. 

From Figure \ref{fig:categoryforward} and Figure \ref{fig:categorybackward}, we notice that: (i) tasks of distinct categories indeed demonstrate different performances for both forward-transfer and backward-transfer evaluations; (ii) the phenomena on the two evaluations are similar: some categories  consistently benefit more, such as ``classification'', ``answer generation'', ``question generation'', while some keep obtaining worse scores, such as ``minimal modification'' and ``verification'' categories. We think this discrepancy origins from two factors; one is how many tasks  a particular category has, the other is how similar or relevant the tasks in that category are with tasks of other categories. Intuitively, a category with more tasks occupying the task chain and resembling other tasks, such as ``classification'', ``answer generation'' and ``question generation'',  can be easier solved when the model comes up to it or comes back to it.

% \paragraph{$\mathcal{Q}_3$: how much does each item in an instruction  contribute?} Since the focus of this work is to solve a new task by understanding its instruction, and each instruction consists of multiple items (please refer to Figure \ref{fig:instruction}), we wonder how necessary each item is. To this end, we discard items \texttt{Title}, \texttt{Definition}, \texttt{Caution}, \texttt{Things to avoid} and \texttt{Prompt} from the instruction separately. As Figure \ref{fig:influenceofitem} shows, although there is no absolute order among the contributions of those instruction items, we can conclude that \texttt{Definition}, \texttt{Prompt} and \texttt{Things to avoid} are always the dominant factors while \texttt{Title} contributes negligibly.

\section{Conclusions}
This work introduced a novel learning problem: continual learning from  task instructions. The goal is to explore the potential of exiting pretrained language models in solving new tasks by understanding instructions rather than labeled examples. With our problem formulation and a well-performing system, we pave the way for future study of this challenge in the community. 

\section*{Acknowledgement}
We thank Daniel Khashabi from AI2 and Swaroop Mishra from ASU for help during this work.   

\bibliography{acl}
\bibliographystyle{acl_natbib}

% \appendix

% \section{Example Appendix}
% \label{sec:appendix}

% This is an appendix.

\end{document}